\documentclass[11pt,oneside,runningheads]{taima}

\usepackage{graphicx, float, epsfig, url, amssymb, tipa}

\begin{document}

\pagestyle{headings}

\mainmatter

\title{
    \Large{Amélioration des Performances des Systèmes Automatiques de Reconnaissance de la Parole pour la Parole Non Native}
}

\titlerunning{Reconnaissance de la parole non native}

\author{\large{Ghazi Bouselmi\inst{1} \and Dominique Fohr\inst{1}  \and Irina Illina\inst{1,2}  \and Jean-Paul Haton\inst{1,3}  }}

\authorrunning{G. Bouselmi, D. Fohr, I. Illina et J.-P. Haton}

\institute{
    Laboratoire Lorrain de Recherche en Informatique et ses Applications - U.M.R. 7503,
    Groupe Parole,\\
    LORIA - Campus Scientifique - BP 239 - 54506 Vandoeuvre-lès-Nancy Cedex, France.\\
    Tél: int+ 33 3 83 59 30 00, Fax: int+ 33 3 83 41 30 79\\
    \email{bousselm@loria.fr, fohr@loria.fr, illina@loria.fr, jph@loria.fr\vspace{1cm}} \\
    \and
    Université Nancy II, IUT Charlemagne,\\
    2 Boulevard Charlemagne - 54000 Nancy, France.\\
    Tél: int+ 33 3 83 91 31 06, Fax: int+ 33 3 83 28 13 33\\
    \email{illina@iuta.univ-nancy2.fr\vspace{1cm}} \\
    \and
    Université Henry Poincaré, Nancy 1,
    Faculté des sciences et techniques,\\
    Campus Victor grignard - BP 239 - 54506 Vandoeuvre-lès-Nancy Cedex, France.\\
    Tél: int+ 33 3 83 68 40 00, Fax: int+ 33 3 83 68 40 01\vspace{1cm}
}

\maketitle %

\begin{abstract}
Dans cet article nous décrivons une approche pour la reconnaissance automatique de la parole (RAP) non native.
Nous proposons deux méthodes pour l'adaptation d'un système de reconnaissance automatique de la parole (SRAP) existant.
La première se base sur la modification des modèles acoustiques par l'intègration des modèles de la langue maternelle (LM).
Les phonèmes de la langue parlée (LP) sont prononcés d'une manière proche des phonèmes de la LM du locuteur.
Nous proposons de ``fusionner'' les modèles de ces phonèmes confondus de manière à ce que le système puisse reconnaitre les deux prononciations concurrentes.
La deuxième approche que nous proposons une détection d'erreurs de prononciation plus précises en introduisant des contraintes graphémiques à ce processus.
En effet, certaines réalisations phonétiques non natives sont fortement guidées par la graphie des mots.
Les améliorations de performances du SRAP adapté selon notre approche sont en moyenne de 22.5\% en taux de phrases correctes et 34.5\% en taux de mots corrects.

\vspace{0.4cm}%
\textbf{Mots clés} Reconnaissance de la parole non native, confusion phonétique, modification de HMM, contraintes graphèmiques.
\end{abstract}

\section{Introduction}
La RAP est de plus en plus utilisée dans plusieurs services grand public tels que les centres d'appels et les centres de billeteries.
Cela est d\^u au développement des technologies utilisées et à la maturation des méthodes de reconnaissance de la parole.
Toutefois, ce type de services peut \^etre confrontés à des utilisateurs dont la langue maternelle n'est pas celle pour laquelle le SRAP a été conçu.
Dans de telles circonstances, le bon fonctionnement du système pourrait \^etre corrompu.
En effet, face à de la parole non native, les performance des SRAP chutent grandement.\\

Les langues parlées utilisent un nombre limité de sons de ``{\it l'espace phonétique} (ou l'espace des sons possibles).
Cet ensemble de sons est spécifique à chaque langage et diffère en nombre de phonèmes d'une langue à une autre.
En effet, pour chaque langue, une partie de l'espace phonétique est partitionnée en un certain nombre de domaines correspondant chacun à un son (ou phonème) particulier.
Les locuteurs d'une certaine langue sont accoutumés à bien reconnaitre les sons propres à leur langage et à les prononcer correctement au sein de ces ``{\it frontières phonétiques}''.
Ce partitionnement de l'espace phonétique est fonction de la langue considérée dans la mesure où seule une partie de cet espace est utilisée.
De ce fait, certains phonèmes d'une langue peuvent ne pas appara\^itre dans une autre.
Un locuteur non natif pourrait remplacer ces phonèmes inexistants dans sa langue maternelle par des sons de cette dernière qu'il considère proches.
Par exemple, selon les travaux de Ladefoged et al.~\cite{Ladefoged} et Jeffers et al.~\cite{RNL}, une grande partie de locuteurs grecs prononcent le phonème {/h/} (présent dans le mot anglais \textit{{\bf h}otel}) comme le son {\textipa{[x]}} (présent dans le mot allemand {\it ba{\bf ch}}).
Les diphtongues anglaises telles que \textipa{[aI]} constitueraient un autre exemple: elle n'existent pas dans la langue française et sont prononcées comme la suite de phonèmes
français \textipa{[a][I]}.
De plus, certains phonèmes très proches (de point de vue acoustique) d'une langue peuvent \^etre assimilés à un m\^eme son dans une autre.
Ainsi les sons du français présents dans \textit{br{\bf in}} et \textit{br{\bf un}} peuvent \^etre confondus par un locuteur non natif et prononcés de la m\^eme manière.\\

Les SRAP sont généralement basés sur des méthodes statistiques et entra\^inés sur des bases de données de parole native.
Par conséquents, les SRAP traditionnels ne gèrent pas les accents non natifs et leurs performances chutent face à ces prononciations exotiques.
L'adaptation des SRAP à la parole non native consiste à augmenter leur tolérance vis-à-vis des ces accents étrangers.
Plusieurs techniques ont déjà été développées pour la reconnaissance de ces accents non natifs.
Elle diffèrent en fonction de la méthodologie adoptée pour la définition des accents non natifs (assimilés à des erreurs de prononciation)
ainsi qu'en fonction des modifications introduites dans le SRAP par la suite.
Les accents étrangers peuvent \^etre définis à travers l'expertise de spécialistes humains comme dans l'approche décrite dans~\cite{ES1}.
Ces erreurs de prononciation peuvent également \^etre détectés automatiquement via une reconnaissance phonétique sur une base de données de parole non-native.
Dans les travaux de~\cite{ES2,ES4}, le SRAP de la LP fournit un alignement phonétique de la prononciation canonique sur la base de données tandis que
le SRAP de la LM est fournit une transcription phonétique (reconnaissance phonétique).
Les deux transcriptions sont ensuite alignées afin de détecter les erreurs de prononcation et établir une matrice de confusion phonétique.\\
Les modifications apportées au SRAP peuvent \^etre appliquées en différents de la structure de ce derniers.
En effet, les approches de~\cite{ES1,ES2} introduisent les différentes prononciations non natives dans le dictionnaire du SRAP.
J. Morgan~\cite{ES4} fusionne les modèles acoustiques des phonèmes de la LP et de la LM qui ont été mis en relation dans la matrice de confusion.\\

Dans cet article nous décrire notre approche pour la reconnaissance de la parole non native introduite dans \cite{ES0,ES00}.
Dans la section~\ref{section_confusion} nous allons aborder la détection des erreurs de prononciation des locuteurs non natifs
ainsi que l'adaptation du SRAP selon ces dernières. Nous aborderons, dans la section~\ref{section_contraintes_graphemiques}, les contraintes graphèmiques dans ces dites erreurs et nous décrirons
notre approche pour leur utilisation. Nous allons ensuite décrire les tests effectués et nous discuterons leurs résulats.
Nous cloturerons enfin avec une brève conclusion.

\section{Erreurs de prononciation}
\label{section_confusion}

Comme expliqué plus haut, les langues parlées utilisent des sons différents.
Lorsqu'un locuteur prononce de la parole étrangère, il ne fait qu'imiter les intonnations et accents de la langue cible.
Il prononce des sons de sa langue maternelle adaptés aux sonorités de la langue parlée.
Dans le cas de figure où certains sons à prononcer n'existent pas dans sa LM, un locuteur peut imiter ce son ou encore le remplacer par un son de sa LM qu'il considère
proche. C'est le cas du son anglais /h/ prononcé par des locuteurs grecs comme le phonème grec \textipa{[x]}, ou encore le son anglais \textipa{[D]} prononcé par certain locuteurs français
comme les phonèmes (français) \textipa{[z]} ou \textipa{[s]}.
De plus, certain phonèmes de la LP peuvent \^etre réalisés comme une suite de la LM comme c'est le cas pour les diphtongues anglaises n'existant pas dans la langue française.\\

Nous avons donc considéré l'utilisation des modèles acoustiques de la LM des locuteurs non natifs dans l'extraction des erreurs de prononciation phonétique (confusion phonétique).
Ainsi, un phonème de la LP peut \^etre mal prononcé comme une ou plusieurs suites de phonèmes de la LM.
Cette confusion phonétique est ensuite utilisée pour la modification du SRAP.
Nous avons conçu une métode pour inclure toutes les prononciations non-natives possibles dans le SRAP tout en concervant l'integrité temporelle des modèles acoutiques
et minimisant les surco\^uts de calculs pour la reconnaissance. Les modèles acoustiques des phonèmes de la LM (HMM: modèles de markov cahés) sont rajoutés comme des chemin alternatifs
dans le modèle du phonème de la LP avec lequel ils ont été confondus.

\begin{figure}[htpb]
\centerline{\epsfig{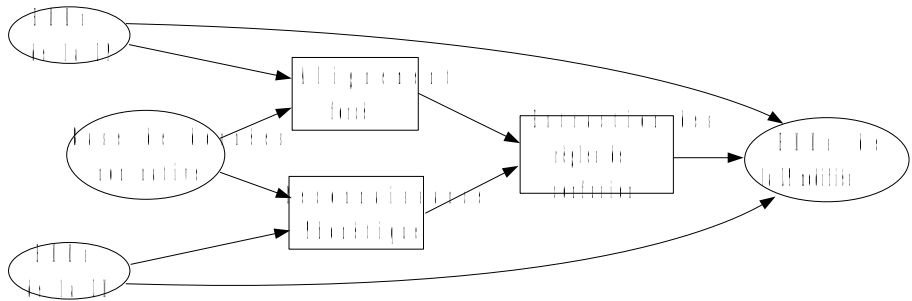}}
\caption{{\it Extraction et utilisation de la confusion phonétique.}}
\label{figure_overview1}
\end{figure}

La figure~\ref{figure_overview1} illustre les processus d'extraction et d'utilisation de la confusion phonétique.
Nous allons décrire ces deux approches plus en profondeur dans les sections suivantes.

\subsection{Détection des erreurs de prononciation}

Afin de déterminer les erreurs de prononciations des locuteurs non natifs d'une certaine origine, nous recourons à une base de données de parole non native étiquetées.
Pour chaque phrase de cette base de données, nous effectuons ce qui suit:
\begin{itemize}
\item un alignement forcé avec le SRAP de la LP, ce qui donne une transcription phonétique en terme de phonèmes de la LP avec leurs intervalles d'occurrence.
\item une reconnaissance phonétique avec le SRAP de la LM, ce qui donne une transcription phonétique en terme de phonèmes de la LM avec leurs intervalles d'occurrence.
\item ces deux transcriptions sont comparée afin d'associer les phonèmes de la LP aux suites de phonèmes de la LM qui se produisent dans le même intervalle de temps.
\end{itemize}

Nous obtenons ainsi, pour chaque phonème $P^P$ de la LP un ensemble d'associations (avec des suites de phonèmes de la LM) avec leurs nombres d'occurrences. Seules les associations les plus fréquentes -pour chaque phonème de la LP- sont retenues pour former l'ensemble des règles de confusion phonétique.

\subsection{Modification des modèles acoustiques}

Afin de prendre en compte de toutes les prononciations alternatives pour chaque phonème $P^P$ de la LP, nous adjoignons au HMM de $P^P$ la concatenation des modèles des phonèmes présents dans les règles de confusion de $P^P$.
En d'autres termes, les modèles HMM des phonèmesde chaque règle de confusion concernant $P^P$ sont concaténés et ajoutés comme chemin alternatif dans le modèle de $P^P$. Ce modèle modifié sera utilisé ultérieurement en lieu et place du modèle canonique de $P^P$ dans le SRAP modifié. De ce fait, le moteur de reconnaissance aura la possibilité de reconnaitre la prononciation canonique de $P^P$ ainsi que toutes ses prononciations alternatives.\\

Pour illustrer ce processus, considérons l'exemple de règles de confusion pour le phonème anglais \textipa{[aI]} où la LM est le français:

\begin{tabular}{ll}
\textipa{[aI]} $\rightarrow$ \textipa{[a][e]}\ \ \ \ \ \ \ \ \  & P(\textipa{[aI]} $\rightarrow$ \textipa{[a][e]}) = 0.4 \\
\textipa{[aI]} $\rightarrow$ \textipa{[a][I]}\ \ \ \ \ \ \ \ \  & P(\textipa{[aI]} $\rightarrow$ \textipa{[a][I]}) = 0.6 \\
\end{tabular}

\begin{figure}[htpb]
\centerline{\epsfig{figure=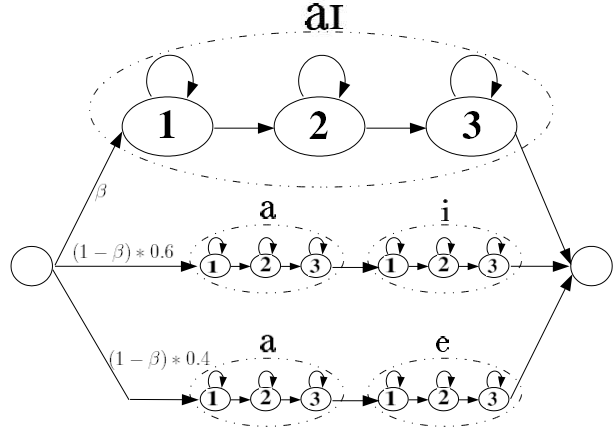,width=80mm}}
\caption{{\it HMM adapté pour le phonème \textipa{[aI]}.}}
\label{figure3}
\end{figure}

Deux chemins additionnels sont rajouté au HMM du phonème \textipa{[aI]} correspondant aux deux règles précédentes: le chemin \textipa{[a][e]} et le chemin \textipa{[a][i]}. Le schémas du modèle adapté est donné dans la figure~\ref{figure3} ($\beta$ est une pondération).

\section{Contraintes graphémiques}
\label{section_contraintes_graphemiques}

L'utilisation des contraintes graphémiques est motivé par l'influence de la graphie des mots sur les erreurs de prononciation non native dans le cadre de la parole lue. En effet, nous avons constaté l'existance d'une relation entre les caractères composant les mots lue et les erreurs phonétiques produites par les locuteurs. Cela pourrait résulter des mécanismes de lectures de la langue maternelle pour des lettres séparées ou encore de la similitudes des constructions graphémiques entre la LP et la LM pour des mots entiers.
Certains locuteurs non natifs pourraient avoir recours aux règles de transformation graphème-phonème de leur LM pour prononcer certains mots ou syllabes.
Par exemple, le mot anglais \textit{minus} (\textipa{[m][aI][n][@][s]}) a été prononcé comme \textipa{[m][i][n][u][s]} par un locuteur italien.
Les mots anglais \textit{approach} (\textipa{[@][p][r][@U][tS]}) 
et \textit{position} (\textipa{[p][@][z][i][S][@][n]}) ont été prononcé comme 
(\textipa{[a][p][r][O][t][S]}) et (\textipa{[p][O][z][i][S][O][n]}) par un locuteur français (respectivement). Dans le dernier exemple, le phonème \textipa{[@]} a été prononcé comme le phonème français \textipa{[a]} (resp. \textipa{[O]}) lorsqu'il correspond au caractère 'a' (resp. 'o').\\

Nous avons donc contraint les règles de confusion décrites plus haut par les caractères auxquels sont reliés les phonèmes.
Pour l'exemple précédent, le phonème \textipa{@} relié au charactère 'a' (resp. 'o') serait confondu avec le phonème \textipa{[a]} (resp. \textipa{[O]}).
Pour ce faire, nous avons adopté une approche basée sur des HMM discrèts (DHMM) afin d'aligner les phonèmes et les graphèmes dans la prononciation de chaque mot du dictionnaire de notre application.
Dans ce système, les états émétteurs représentent les phonèmes et les observations discrètes représentent les caractères.
Ce système DHMM est entrainé sur un grand dictionnaire phonétique.

\section{Expérimentations et résutats}

\subsection{Conditions expérimentatles}

Notre travail se situe dans le cadre d'un projet européen visant le développement d'un système d'aide pour les pilotes d'avions basé sur des commandes vocales.
Pour nos tests, nous disposons d'une base de données de parole non native composée 31 locuteurs français, 20 grecs, 20 italiens et 10 espagnols.
Chacun de ses locuteurs a prononcé 100 pharses en langue anglaise.
Ces phrases suivent une grammaire de commande stricte.
Le vocabulaire est composé de 134 mots.
Nous avons utilisé le système de reconnaissance HTK pour nos expérimentations.
Les modèles acoustiques sont des HMM à 3 états gauche-droite avec des mixture de  128 Gaussiennes.

\subsection{Tests et résulats}

Nous avons adopté la méthode de validation croisée dans nos tests (cross-validation, leave-one-out). Afin de terter un locuteur donnée d'une origine $X$, le reste des locuteurs originaires de $X$ sont utilisés pour l'extraction des règles de confusion. Nous avons effectué des tests avec une grammaire stricte et une grammaire libre (boucle de mots). Nous avons également testé l'adaptation au locuteur avec la technique MLLR (Maximum Likelihood Linear Regression). Le tableau~\ref{table1} résume les résultats obtenus.
La confusion phonétique a permis une réduction du taux d'erreur de phrases (sentence error rate, SER) variant de 9\% à 48\% (relatif) et une réduction du taux d'erreur de mots (word error rate, WER) allant de 19\% à 46\% (relatif), et ce en utilisant la grammaire contrainte. Avec la grammaire libe, la réduction du taux SER varie de 6\% à 49\% et le taux WER est diminué d'une proportion allant de 20\% à 33\%. En moyenne, les taux SER et WER sont réduits de 34.5\% et 22.5\%.\\

Il est intéressant de noter que l'utilisation des contraintes graphémiques a permis une meilleure perfromance du système avec la grammaire libre.
Il est également important de noter que la confusion phonétique affiche des performances proches, sinon meilleures, que l'application de l'adaptation au locuteur (MLLR).
Ceci est d'autant plus frappant dans le test de la grammaire libre où l'on observer une différence relative moyenne de 23\% sur le taux WER et 10\% sur le taux SER.

\begin{table*} [ht]
\caption{\label{table1} {\it Résultats de test pour les bases de données Francaises, Italienne, Espagnole et Grèque (en \%).}}
\centerline{
\begin{tabular}{|l|c|c|c|c|c|c|c|c|c|c|}
\hline
        & \multicolumn{2}{|c|}{\it Francais} & \multicolumn{2}{|c|}{\it Italien} & \multicolumn{2}{|c|}{\it Espagnol} & \multicolumn{2}{|c|}{\it Grec} & \multicolumn{2}{|c|}{\it Moyenne} \\ \hline
\textit{System} & \textit{WER} & \textit{SER} & \textit{WER} & \textit{SER} & \textit{WER} & \textit{SER} & \textit{WER} & \textit{SER} & \textit{WER} & \textit{SER}\\
\hline  \hline
\textbf{\it grammaire stricte:} && &&&& && &&\\
- \textit{baseline} & 6.0 & 12.8 & 10.5 & 19.6 & 7.0 & 14.9 & 5.8 & 13.2& 7.3 & 15.1\\
- \textit{``confusion phonétqiue''} & {4.9} & {11.7} & {6.3} & \textbf{13.0} & {4.8} & {10.9}& \textbf{3.0} & \textbf{7.8}& \textbf{4.8} & \textbf{10.8}\\
\begin{minipage}{3.8cm}
{- \it ``confusion phonétqiue'' +
  contraintes graphèmiques}
  \end{minipage} & \textbf{4.5} & \textbf{11.1} & \textbf{6.2} & {13.2}& \textbf{4.6} & \textbf{10.3} & 4.5 & 11.1& {5.0} & {11.4}\\
\hline
- \textit{baseline + MLLR} & 4.3 & 8.9 & 7.3 & 13.6 & 5.1 & 11.1 & 3.6 & 9.4 & 5.1 & 10.8\\
- \textit{``confusion phonétqiue'' + MLLR} & {3.2} & {7.6} & \textbf{4.9} & \textbf{11.4} & {3.2} & {7.8}& \textbf{2.1} & \textbf{6.1}& \textbf{3.3} & \textbf{8.2}\\
\begin{minipage}{3.8cm}
{- \it ``confusion phonétqiue'' +
  constraintes graph. + MLLR}
  \end{minipage} & \textbf{3.0} & \textbf{7.3} & 5.1 & 11.7 & \textbf{2.8} & \textbf{6.0} &  2.5 & 6.8 & {3.4} & {8.0}\\
\hline \hline
\textbf{\it grammaire libre:} && &&&& && &&\\
- \textit{baseline} &  37.7 & 47.9 & 45.5 & 52.0 & 39.9 & 53.5 & 36.7 & 40.0 & 40.0 & 50.7 \\
- \textit{``confusion phonétqiue''} & {27.8} & {42.6} & {28.9} & {44.9} & {27.1} & {43.0} & {20.1} & {34.0}& {26.0} & {41.1}\\
\begin{minipage}{3.8cm}
{- \it ``confusion phonétqiue'' +
  contraintes graphèmiques}
  \end{minipage} & \textbf{25.6} & \textbf{41.0} & \textbf{26.7} & \textbf{43.2} & \textbf{25.2} & \textbf{40.8}  & \textbf{18.6} & \textbf{33.0}& \textbf{24.0} & \textbf{39.5}\\
\hline
- \textit{baseline + MLLR} & 28.4 & 39.4 & 34.9 & 46.5 & 32.3 & 48.3 & 28.5 & 41.0& 32.2 & 42.7\\
- \textit{``confusion phonétqiue'' + MLLR} & {22.9} & {37.0} & {24.7} & {40.9} & {24.0} & {39.5}& {17.1} & {30.1}& {22.2} & {36.9}\\
\begin{minipage}{3.8cm}
{- \it ``confusion phonétqiue'' +
  contraintes graph. + MLLR}
  \end{minipage} &  \textbf{20.9} & \textbf{35.0} & \textbf{22.7} & \textbf{39.3} & \textbf{21.0} & \textbf{36.3} & \textbf{16.0} & \textbf{29.2} & \textbf{20.1} & \textbf{35.0}\\
\hline
\end{tabular}
}
\end{table*}

\section{Conslusion}

Dans cet article, nous avons décrit notre méthode d'adaptation de système automatique de reconnaissance vocale vis-à-vis de la parole non native.
La confusion phonétique que nous avons mis au point associe à chaque phonème de la langue parlée un ensemble de suites de phonèmes de la langue maternelle.
Chacune de ces suites représente une prononciation alternative spécifique à la langue maternelle traitée.
Nous avons également présenté une approche originale pour inclure ces prononciations sans perte de performances de calcul.
Nous avons enfin décrit une méthode pour l'adjonction de contraintes graphèmiques à la confusion phonétique.

\section{Remerciements}

Notre travail a été partiellement financé par le projet européen \textit{HIWIRE} (\emph{Human Input that Works In Real Environments}), contract numéro 507943,
\emph{sixth framework program, information society technologies}.

\bibliographystyle{plain}
\bibliography{./taima07}

\end{document}